\documentclass{article}

\usepackage{amssymb}
\usepackage{amsmath}
\usepackage[dvips]{graphicx}

\begin{document}


\title{The Harmonic Theory \\ A mathematical framework to build intelligent contextual and adaptive computing, cognition and sensory system }
\author{Nick Mehrdad Loghmani}
\renewcommand{\today}{January 17, 2011}
\maketitle

\noindent 
\newline
\noindent  \textbf{Background:} \newline
Most contemporary computing systems are based on Von Neumann architecture\footnote{(Blundell, 2008)(2008). Computer Hardware. Cengage Learning EMEA. p. 54. ISBN 9781844807512.}. While such systems are efficient in solving certain mathematical problems involving data storage, retrieval and repetitive and predetermined tasks, they are less successful in operating outside of the well-defined envelope they are designed for. Furthermore, interaction with such systems has to happen in a very specific and narrow spectrum of interfaces, with limited margin of flexibility and adaptability.

\noindent 
\newline
\noindent \textbf{BRIEF SUMMARY}
\newline
\noindent Harmonic theory provides a mathematical framework to describe the structure, behavior, evolution and emergence of harmonic systems. A harmonic system is context aware, contains elements that manifest characteristics either collaboratively or independently according to system's expression and can interact with its environment.  This theory provides a fresh way to analyze emergence and collaboration of ``ad-hoc'' and complex systems.

\noindent \newline \textbf{Harmonic system\footnote{``Harmonic System'' is an intelligent contextual computing, cognition and sensory system operating based on the principle of the Harmonic theory.} component, characteristics and relation model explained}

\begin{enumerate}
\item \textbf{ }A harmonic system is defined within an environment. Environment provides the boundary for the system and resources necessary for forming and keeping compositions. The environment also acts as a medium for exchange with other systems. Environment's characteristic model influences system's expression and context as well arrangement and type of characteristics its composition's manifests. Diversity in environment's characteristic model ensures sustainability of harmonic state. Trade and exchange, along with transformation, are among the factors that influence and transform the harmonic environment.

\item  A Harmonic system has a context. Context is the situational constraint set by the system's structure (environment, composition, expression) as well as by its action (state, evolution, transformation, exchange). Harmonic state depends on the harmonic value of compositions relative to harmonic expression. Role, granularity and arrangement of characteristics a composition manifests depend on the context it is in.  By imposing patterns, restrictions and control of the domain of harmonic compositions context controls measurability and comparability of characteristic model.

\item  A Harmonic system has a \textit{Harmonic expression.} Harmonic expression creates the criteria and characteristic model based on which (in a given context) harmonic compositions can be arranged and rated. 

\item  A Harmonic system contains \textit{compositions.} Composition is a pattern of entities from the harmonic environment formed to manifest a range of characteristics in accordance to expression and context.
\end{enumerate}

\noindent There are three types of compositions:

\begin{enumerate}
\item \begin{enumerate}
\item \begin{enumerate}
\item  Active composition is present in the harmonic system environment and contributes to harmonic state\footnote{Herein after referred to as having a harmonic relation with the harmonic expression}

\item  Passive composition is present in the harmonic system environment but does not have a direct relation with harmonic (Weisstein E. W.) (Weisstein E. W.) (Weisstein E. W.)expression. However it has a characteristic model, which is in demand by other systems. Therefore it can be used as a resource in an exchange scenario. 

\item  Target composition is NOT present in the harmonic system environment. However, if obtained,(either through exchange with other systems, through transformation, or through a combination of both) it can boost the system's harmonic status.
\end{enumerate}
\end{enumerate}
\end{enumerate}

\noindent 

\noindent As a general rule any harmonic system with a suitable characteristic model can be used as a composition in a larger system. There are two categories of characteristic model; referential or intrinsic; intrinsic characteristic are those which stem from composition's structure and natural makeup, for instance water molecular characteristic model is H2O, referential characteristics are those which are observed or infer by a third party system and as such are subjective and relative to composition real characteristic properties, for instance ``Water'' is the referential characteristic of H2O in the context of English language. 

\noindent Characteristics role can be classified as:

\noindent Inhibitor: reduces the harmonic value of specific characteristics by suppressing or compensating their manifestation to reduce system risk (if characteristics are risk factors) or promotes a competing characteristics. Inhibitors play an important role, as they limit the scope of available options and allow the system to focus its resources

\noindent Activator: increases the harmonic state by directly conforming to harmonic expression. A pattern of characteristics is activator of a certain harmonic pattern if its conformance to the harmonic expression causes the expansion of the latter.

\noindent Facilitator: increases the harmonic state indirectly by enhancing and amplifying contributing characteristics

\begin{enumerate}
\item  System components, including environment, context, expression and composition, all have Abstract Characteristic Models. A Characteristic Model is the derivative of the underlying subject (environment, context, composition or expression). As mentioned earlier, depending on the context, a characteristic model represents either the stable or dynamic aspect of the underlying subject. The evolution and transformation of a subject influence its characteristics. 
\end{enumerate}

\noindent 

\noindent Harmonic value provides a method to compare a manifested characteristic\footnote{Characteristic of the underlying subject} to the expressed or desired characteristic; Harmonic value is calculated by comparing the characteristics of H${}_{c}$ and H${}_{e}$ as represented on the Harmonic helix\footnote{Harmonic characteristics for H$  $c$  $ \& H$  $e$  $ are plotted as points on a double helix structure circumference, its radius represents observation distance and arc length between observed and reference characteristic represents measure of comparison} by c${}_{1}$ to c${}_{n}$ and e${}_{1}$ to e${}_{n}$ respectively; where L${}_{obs}$=measure of comparison, is the observed difference between two characteristics and  L${}_{t}$ = measure of comparison, is the true difference in a given context between two characteristic as measured on an intertwined double helix where:

\begin{enumerate}
\item  $\theta $ is the angle between observed and reference characteristic

\item  $\mu $ is the magnitude of observed characteristic measured by its deference from reference characteristic divided by $\pi $: \includegraphics*[width=0.46in, height=0.42in, keepaspectratio=false]{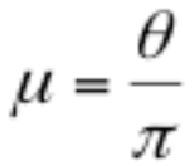}

\item  Its radius is 1 (for simplicity for single context can be represented by a unit circle\footnote{ $  Weisstein, Eric W., http://mathworld.wolfram.com/UnitCircle.html$ }) 

\item  \textit{r} is the distance of observation (radius of comparison circle); for the unit circle  r=1
\end{enumerate}

\noindent \includegraphics*[width=3.44in, height=2.85in, keepaspectratio=false]{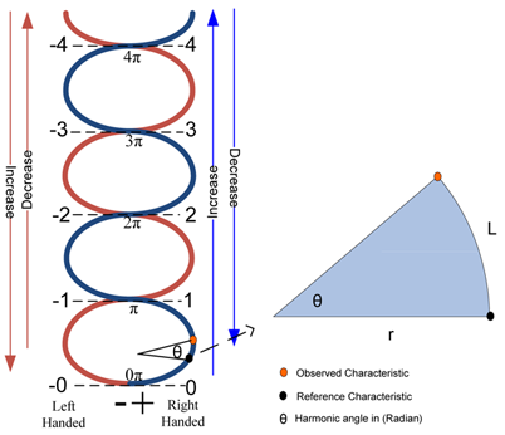}

\noindent \textbf{Figure 1 Harmonic value is calculated by comparing the characteristics of H${}_{c}$ and H${}_{e}$ as represented on the Harmonic helix\footnote{Harmonic characteristics for H$  $c$  $ \& H$  $e$  $ are plotted as points on a double helix structure circumference, its radius represents observation distance and arc length between observed and reference characteristic represents measure of comparison} by c${}_{1}$ to c${}_{n}$ and e${}_{1}$ to e${}_{n}$ respectively; where L${}_{obs}$=measure of comparison}

\noindent 

\noindent Harmonic value is calculated using the following formula:\includegraphics*[width=1.34in, height=0.51in, keepaspectratio=false]{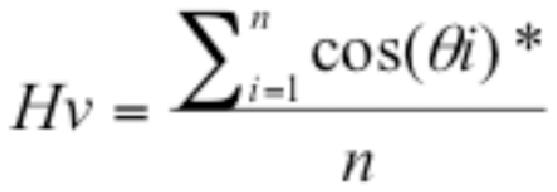}\includegraphics*[width=1.34in, height=0.51in, keepaspectratio=false]{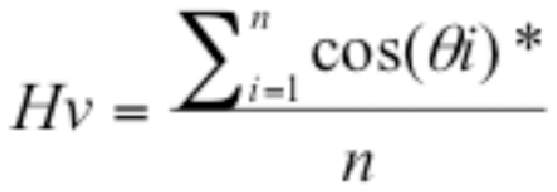}

\noindent 

\noindent  \includegraphics*[width=2.68in, height=2.86in, keepaspectratio=false]{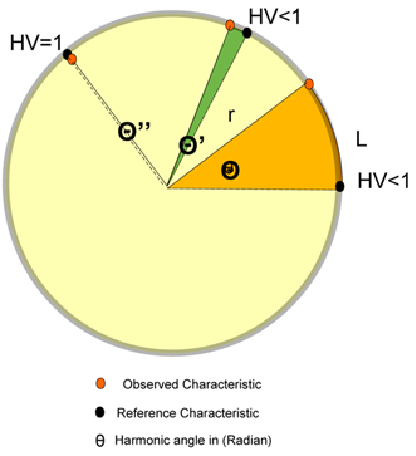}

\noindent \textbf{Figure 2 Harmonic value calculation}

\noindent Where $\theta $ is the angle between characteristic for H${}_{e}$ and H${}_{c}$ expressed in radian\footnote{ $  Max $  $\theta $  $ for two characteristics in the same context$  $-\frac{\pi }{4}<\varphi <\frac{\pi }{4}\ is\ \pi  $ }, as plotted on harmonic helix, with -1 and   1 representing the min and maximum of the H${}_{v}$, and n is the quantity of characteristics for H${}_{c}$ , if $\theta $=0 then characteristic has harmonic value of 1. 

\noindent 

\begin{enumerate}
\item  The underlying subject significance is highlighted by its \textit{Harmonic Significance.} Harmonic Significance is a weighted harmonic value:
\end{enumerate}

\noindent \includegraphics*[width=1.33in, height=0.51in, keepaspectratio=false]{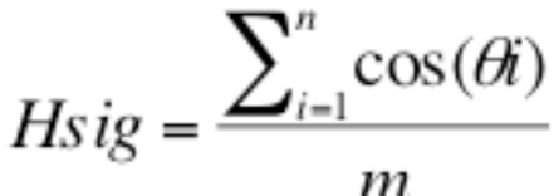}

\noindent 

\noindent Where \textit{m} represents the total number of characteristics belonging to all compositions participating in the expression characteristic model\footnote{Depending on the context, expression can have multiple characteristic model}.

\noindent Harmonic Significance indicates the harmonic composition potential for contributing to the harmonic state. Significance is represented by the granularity of composition characteristics, with more significant compositions represented by a higher number of characteristics. In a given context if multiple compositions have equal or close harmonic value, harmonic significance become a deciding factor for composition selection.

\noindent 

\begin{tabular}{|p{1.4in}|p{1.1in}|p{1.1in}|} \hline 
Significance & Expression Characteristics & Composition Characteristics \\ \hline 
Increase & Increase & Increase \\ \hline 
Decrease & Decrease & Don't care \\ \hline 
\end{tabular}

\noindent Significance increases when the granularity of `expression characteristics' \textit{\underbar{and}} conforming\footnote{Conforming characteristic have positive HV} `composition characteristics' increases\footnote{Increasing granularity of characteristic for expression and composition is called enrichment}; Significance decreases when the granularity of `expression characteristics' decreases\footnote{Decreasing granularity of characteristic for expression and composition is called simplification}.

\begin{enumerate}
\item  \textit{Harmonic state} represents the quadratic mean\footnote{Kenneth V (Fall 2007), "Determining the Effective or RMS Voltage of Various Waveforms without Calculus", Technology Interface 8 (1): 20 pages} (RMS) of harmonic values of the selected\footnote{Selected composition are those which have the highest significance} compositions (selected based on their significant) in a given context. Harmonic State (H${}_{s}$) is calculated by the following formula:
\end{enumerate}

\noindent \includegraphics*[width=1.49in, height=0.46in, keepaspectratio=false]{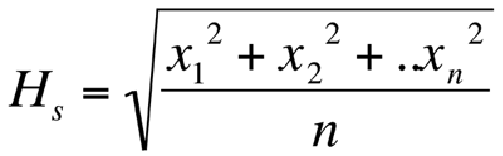} 

\noindent where \{X${}_{1}$,X${}_{2,..}$X${}_{n}$\} represent harmonic values of the selected compositions in the context and\textit{ }n\textit{ }represents number of selected compositions in the context.\textit{}

\noindent 

\noindent Compositions with higher significance have more positive influence on harmonic state.

\noindent Change in the composition characteristics model can increase or decrease the harmonic significance, therefore changing the system's state. Change in the system's environment can also force a state transition by changing context and compositions.

\noindent The response of a Harmonic system to state transition can be classified as: 

\noindent \begin{enumerate}
\item á áReactive; system responds to change reactively, after state transition has been detected, either by enriching composition characteristics, to conform to the enriched expression model, or simplifying the expression model, to conform to the simplified composition model.

\noindent \item á áActive; system tries to maintain or achieve a sustainable harmonic state by creating opportunities and avoiding risks preemptively, through enrichment, simplification, exchange or a combination of all three. Active systems retain transformation patterns, which can be reused in similar contexts and subsequently can be replicated in successive evolution, therefore forming a generational memory.
\end{enumerate}

\noindent \textit{}

\begin{enumerate}
\item \textit{ }Harmonic Status (HS): represents the average of harmonic values of all participating compositions over an interval [a,b] where `a' and `b' represent evolution boundaries. 
\end{enumerate}

\noindent 

\begin{enumerate}
\item  Harmonic Exchange Value (X${}_{v}$); harmonic exchange value is the total ''value realized'' by exchanges between systems and depends on the harmonic value realized by each system, divided by the number of exchanging parties.
\end{enumerate}

\noindent \textbf{\textit{Direct Exchange Value}}\footnote{For both parties, each party can calculate its return by using $  $\includegraphics*[width=1.08in, height=0.39in, keepaspectratio=false]{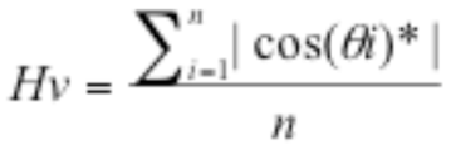}$  $; for instance in a game of chess each participant goal is to maximize own return while minimizing her/his opponent's} between two systems (S${}_{1}$ and S${}_{2}$) is calculated by the following formula:

\noindent \includegraphics*[width=4.24in, height=0.42in, keepaspectratio=false]{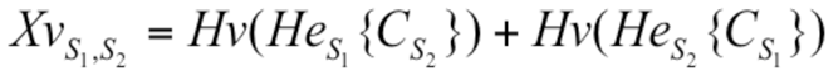}

\noindent \includegraphics*[width=0.35in, height=0.24in, keepaspectratio=false]{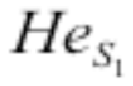} = Harmonic expression of system S${}_{1 }$

\noindent \includegraphics*[width=0.24in, height=0.24in, keepaspectratio=false]{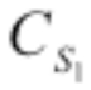}  = Composition of system S${}_{1 }$

\noindent \includegraphics*[width=0.37in, height=0.25in, keepaspectratio=false]{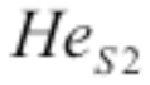}= Harmonic expression of system S${}_{2}$

\noindent \includegraphics*[width=0.26in, height=0.25in, keepaspectratio=false]{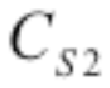}  = Composition of system S${}_{2 }$

\noindent \newline\newline \textbf{\textit{Exchange efficiency}} is:

\noindent \includegraphics*[width=0.81in, height=0.55in, keepaspectratio=false]{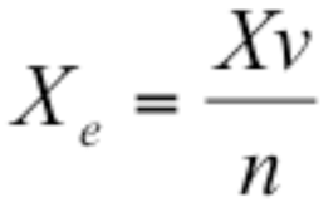} 

\noindent Where \textit{n} is number of exchanging parties in the exchange chain

\noindent 

\noindent The exchange motivation factor is the difference between harmonic state before and after the exchange.

\noindent Longer exchange chains can provide potential value otherwise not attainable in direct exchange. For instance in the following exchange scenario:

\noindent 

\noindent Set A contain: \{(10,12,14),(1,4,5)\}

\noindent Set B contain: \{(2,3),(23,24,25)\}

\noindent Set C contain: \{(6,7,8,9,10),(16,26,27)\}

\noindent 

\noindent  \includegraphics*[width=5.05in, height=1.37in, keepaspectratio=false]{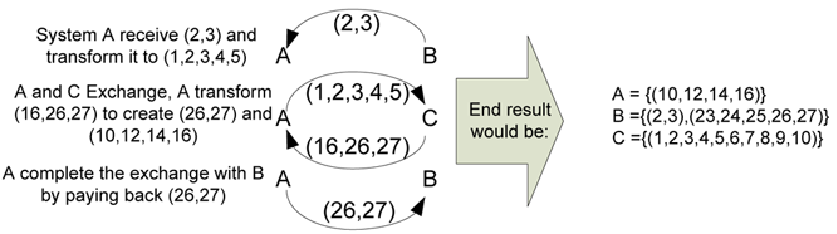}

\noindent \textbf{Figure 3 Composition transformation enhance exchange opportunity}

\noindent System A transforms composition to create an exchange opportunity with B and C\footnote{System A is able to acquire composition from B by differing its payment after trading with C}; the above example is a multi party exchange scenario which involve more than two exchange parties; the length of the exchange chain depends on capability of participating systems to transform and create desirable compositions for further trade and exchange.\textbf{\textit{}}

\noindent \textbf{\textit{Indirect exchange value}} is when payout to one party in the exchange is done through a third party (indirectly)

\noindent \includegraphics*[width=4.50in, height=1.03in, keepaspectratio=false]{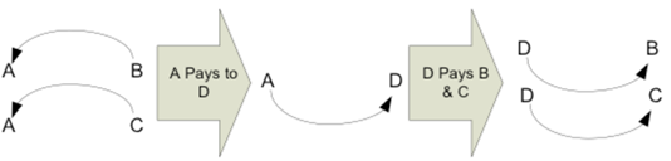}

\noindent \textbf{Figure 4 indirect exchange scenarios}

\noindent An example of the above scenario is the natural recycling process, where composition used by plants and animals is recycled, thus creating a \textbf{\textit{Self sufficient exchange ecosystem}} where exchange chain is self sustaining.

\noindent\includegraphics*[width=5.47in, height=1.38in, keepaspectratio=false]{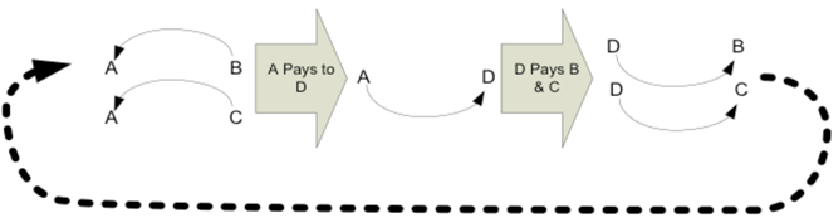}

\noindent \textbf{Figure 5 self-sustaining exchange chain}

\noindent 

\noindent 

\begin{enumerate}
\item  Harmonic Transformation; is change in the characteristics of composition, environment, context or expression to maintain or increase harmonic state, status and exchangeability; there are two types of Harmonic transformation:

\begin{enumerate}
\item  Simplification: 

\begin{enumerate}
\item  Simplifying the underlying subject\footnote{Characteristic model may be maintained or simplified as well} OR

\item  Reducing granularity of the characteristics model\footnote{Underlying subject may be maintained or simplified as well} OR

\item  A combination of both
\end{enumerate}
\end{enumerate}
\end{enumerate}

\noindent Decomposing a composition to create tradable compositions is an example of simplification.

\noindent \includegraphics*[width=1.57in, height=0.25in, keepaspectratio=false]{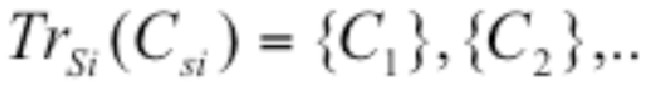}

\begin{enumerate}
\item \begin{enumerate}
\item  Enrichment:

\begin{enumerate}
\item  Increasing granularity of the characteristic model\footnote{While maintaining the underlying subject} 

\item  Enriching the underlying subject and creating potential for increasing granularity of the characteristic model
\end{enumerate}
\end{enumerate}
\end{enumerate}

\noindent Combining compositions to create a tradable characteristic model is an example of Enrichment

\noindent \includegraphics*[width=1.81in, height=0.25in, keepaspectratio=false]{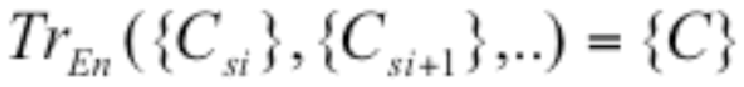}

\noindent Positive transformation patterns emerge when:

\noindent For Context \textit{C${}_{1}$} and Expression \textit{E${}_{1}$} transforming\footnote{Transformation can be either enrichment or simplification} the underlying subject or the characteristic model \textbf{\textit{always}} results in harmonic state being maintained or improved. In other words:

\noindent If composition \textit{A} is exchanged with, or transformed to composition \textit{B},

\noindent \textbf{\textit{OR}}

\noindent If characteristic model \textit{X${}_{1}$}${}_{ }$is transformed to \textit{X${}_{2}$} 

\noindent \textbf{\textit{THEN ALWAYS}}

\noindent Harmonic state S${}_{1 }$is maintained or improved.

\begin{enumerate}
\item  Mathematical and Logical operand in a harmonic system
\end{enumerate}

\noindent Basic Mathematical operation is conducted using the following technique:

\noindent There are two intertwined helix, blue positive, red negative

\noindent \includegraphics*[width=1.23in, height=2.08in, keepaspectratio=false]{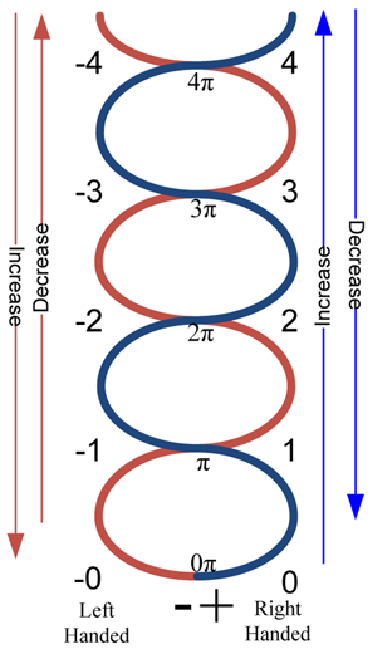}

\noindent \textbf{Figure 6 Math operation using harmonic helix}

\noindent Addition implies combining

\noindent Multiplying implies scaling

\noindent Adding implies increasing

\noindent Deduction implies decreasing

\noindent Increase of increase= increase

\noindent Decrease of increase= decrease

\noindent Decrease of decrease= decrease

\noindent Decrease from positive number is always turn CCW, decrease from negative numbers is always CCW

\noindent Increase from positive number is always turn CW, Increase from negative numbers is always CW

\noindent Helix start at positive 0, turn by 0$\pi $ is negative 0 

\noindent When a number is decreased to zero, further decrease from zero follows negative number rule, start from -0 (decrease from 0 to -0 is 0$\pi $  )

\noindent X+Y= Turn CW by magnitude of Y$\pi $ from X 

\noindent X-Y= Turn CCW by magnitude of Y$\pi $ from X 

\noindent X*Y= if both side increase or decrease:

\noindent Turn CW from 0 magnitude of Y$\pi $, X times Else Turn CCW from 0 magnitude of Y$\pi $, X times

\noindent 
\section{Logical relation}

\noindent 
\section{A AND B = C}

\noindent Expansion of C, the outcome, depends on presence of both Characteristic A and B, in other word A and B are activator for C, if any of them is missing inhibitor prevent expansion of C; condition of existence could be concurrent or delayed (within a defined interval); expansion of C in this case is serial, depending on occurrence and conformance of all characteristics.

\noindent  

\noindent If Outcome of the above equation is equal or less than 0 then the argument is false; otherwise it is true.

\noindent \includegraphics*[width=1.30in, height=1.47in, keepaspectratio=false]{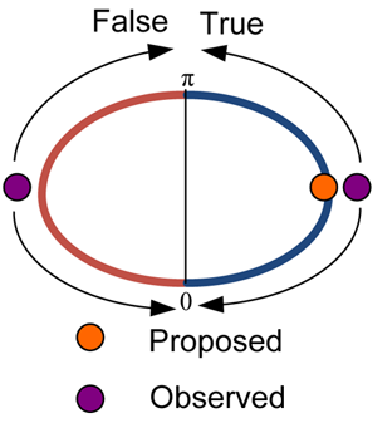}

\noindent \textbf{Figure 7 Boolean operation}

\noindent on Harmonic helix (where positive and negative helix intersect at 0,  $\pi $ ) logical evaluation is represented by transforming predicates on an evaluation helix with every predicate symbol centered on $\pi $/2,  observed $\theta $$>$=90 implies move to negative (Red) helix which is interpreted as false

\noindent \textbf{Forced expansion}

\noindent By injecting enough activator system can be forced into expansion even if one of the operand is not present and inhibits the expansion, effectively neutralizing the effect of inhibitor.

\noindent 

\noindent 
\section{A OR B = C}

\noindent Expansion of C (the outcome) depends on existence of either Characteristic of A or B, unlike conjunction there is no inhibitor, therefore lack of either A or B does not prevent expansion of C. Condition of existence could be concurrent or delayed (within an interval) ; expansion of C this case is parallel, as soon as conformance of any observed characteristic is confirmed expansion of C is triggered.

\noindent Evidently it is slower to examine arguments involving conjunction; therefore there is an opportunity for optimization by transforming cluster of characteristic models into a unit of characteristic model, which can be tested in its entirety.

\noindent \textbf{How recognition happens, Harmonic sensory system}

\noindent Based on harmonic theory principles and in order to focus on desired characteristics and eliminate noise, sensory system is closely being influenced by the context; this process is called sensory system priming.

\noindent \includegraphics*[width=5.89in, height=1.64in, keepaspectratio=false]{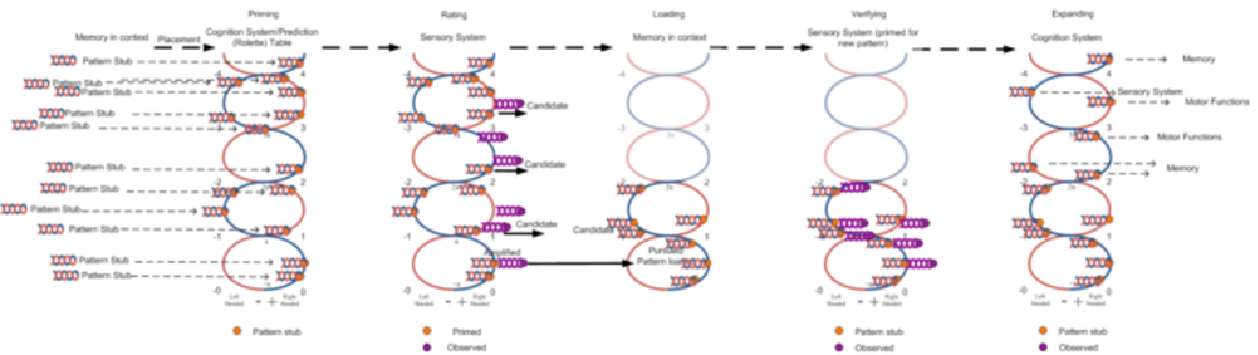}

\noindent \textbf{Figure 8 Harmonic system primed by memory system}

\noindent Above diagram illustrates how sensory system in primed by its relevant memory system (visual, auditory,..); harmonic context determines pattern stub which act as a primer for the sensory system. Harmonic expansion of stub happens when a close enough observed pattern is detected (conformity is evaluated using harmonic value formula); harmonic expansion triggers further cycle of priming and observing.  Optimum cycle frequency is calculated by dividing subject characteristics (represented by observable patterns) divided by quadratic mean\footnote{Chris C. Bissell and David A. Chapman (1992). Digital signal transmission (2nd ed.). Cambridge University Press. p. 64. ISBN 9780521425575.} of cognition system available capacity (represented by primed pattern stub) and sensory system capability (represented by observed pattern), rounded to next whole number.
\[f=\frac{C_{sbj} }{RMS(C_{s} ,C_{C} )} \] 
Optimum frequency is achieved when both cognition system capacity and sensory system capability converge to quadratic mean.

\noindent Cycle's frequency can be further optimized by reducing granularity of subject characteristic model (simplification), or by increasing primed \& observed patterns (capacity of cognitive and sensory system).

\noindent Context has a great influence on priming the sensory system with patterns to look for and selection of patterns to expand.

\noindent \textbf{Thinking }

\noindent Thinking is the ability to use referential characteristics to create scenarios to transform, rearrange and combine various characteristic patterns in order to facilitates emergence of new expression and composition; it also entails evaluating harmonic value and composition arrangement in various context scenarios. This process creates the possibility to observe collaborative emergence of expression and discover cause and effect.

\noindent \textbf{Imagination }

\noindent Imagination is the ability to virtualize sensory system to simulate and invoke referential characteristics (patterns) ``as if they were observed'' to trigger detection and expansion of stub patterns; in other word, arranging patterns and priming them in virtualized sensory system to create real or virtual (imaginative) response\footnote{Virtual response triggers cognitive pattern and limited external motor functions}.

\noindent \textbf{Creativity}

\noindent Creativity is systemic use of referential characteristic, virtualization and transformation to increase efficiency (by reducing cycles) and number of desirable characteristics (through arrangement of composition) as well as the ability to permute and combine compositions to facilitate emergence of new expressions and systems\footnote{New composition arrangement and characteristic model leads to emergence of governing expression} and observe and memorize the process.

\noindent \textbf{Instincts and ``need''}

\noindent Instinct expressions are major contributor to the harmonic system's health and harmonic status, they are required to maintain the integrity, health and continuity of the system, harmonic status that stem from instincts acts as a motivator to maintain and increase compositions which are required to satisfy instinctive expressions\footnote{``Need'' arise from the fact that to maintain the harmonic status there must be enough compositions to provide the required characteristic models} (by manifesting conforming characteristics).

\noindent \textbf{Sensory system and consciousness}

\noindent Harmonic sensory system captures information on its own term\footnote{Harmonic system Sensory system is always being primed by cognitive system}, cognitive system primes sensory system for ``familiar'' or pattern of ``interest''. Sensory system transmits information as ``harmonic patterns'' (which are ``referential'' characteristics); ability to observe and memorize these patterns constitutes  ``\textit{sensational familiarity}''. Sensational familiarity is system's ability to observe, memorize and prime (reproduce) referential characteristics experienced by its sensory apparatus. Sensational familiarity in conjunction with cognitive system ability to project\footnote{In order to create simulation scenarios} referential characteristics of sensory system in ``virtualized sensory'' is the bases for awareness and self-image.

\noindent 

\noindent \textbf{Harmonic System Self awareness and vision of self (Personality)}

\noindent ``Self'' is a Cluster of virtual compositions based on referential and intrinsic characteristics\footnote{Referential and intrinsic characteristics are the result of sensory and cognitive system operation and interactions}, which exhibits\footnote{Referential characteristic help forming a composition with intrinsic characteristics} characteristics pattern and traits according to virtual or real scenario\footnote{Creating a Scenario is done by priming the virtual sensory with stub pattern to trigger a response from cognitive system}. Awareness of ``Self'' as a construct, stems from the ability to observe and predict manifestation\footnote{Manifested capabilities are those patterns which have been observed, memorized and expanded} of characteristics according to stimulus (either virtual or real). Development of self is vital for simulation and imagination, through which system can create and rate permutation of various scenarios to increase efficiency and reduce risk.

\noindent \textbf{Social behavior}

\noindent Social interactions arise from the need to enrich available compositions in the environment, enhance exchange opportunities and increase efficiency by specialization. Collaboration and social interaction aids emergence\footnote{Exchange opportunities to improve harmonic status and exploit opportunity in arrangement of composition to cause system emergence} of new opportunities and reduce risk\footnote{Reduce risk through collaboration to cause emergence of characteristics which are risk inhibitors}; harmonic exchange requires a collaborative and cooperative environment where new composition arrangements would lead to increase opportunity for all of its participants. 

\noindent \textbf{Belief, value system and Morality }

\noindent Risk mitigation, increased social \& environmental interaction and exchange opportunities are among the driving forces to form Harmonic System belief and value system, which is key and influential in its governance. Beliefs and value system comprise of cluster of virtual compositions; which are formed based on observed (or inferred) intrinsic characteristics when manifested by external or internal\footnote{Internal sensational or cognitive composition stemming from instincts provides important input into belief and value system.} compositions. For example socialization increases the odd of survival and increase exchange opportunities; therefore intrinsic characteristic of socialization\footnote{Manifested as cognitive and sensational patterns, for instance exchange capability} contributes to forming system's value \& belief composition.  Intrinsic characteristics based on values and beliefs are strong contributors to harmonic status as they influence the decision to pursuit and expand a pattern stub; if pursuit of pattern causes reduced harmonic value for ``values and beliefs'' characteristics, harmonic status declines substantially\footnote{Degree of decline depends on granularity of ``Values and Beliefs'' system characteristics and granularity and significance of opportunity pattern}.

\noindent \textbf{Emotions }

\noindent Emotions are intrinsic characteristic comprising of sensational and cognitive patterns expressed by instinctive compositions. Demonstration of characteristics by Instinctive compositions depends on context, environment, and observation or detection  (a particular) \textit{arrangement\footnote{Arrangement of characteristics is key in triggering emotional response, $  $arrangement$  $ $  $expression$  $ of particular characteristics can have direct relation to emotional expression}} of characteristics. When manifested, instinctive composition characteristics can lead to expansion of harmonic patterns, expression of referential characteristics (that can be inferred as emotional response) or transformation of context and environment characteristic pattern (which is more subtle). 

\noindent Change in harmonic state, (which itself is governed by intrinsic and referential characteristic conformity to an expressed harmonic expression) can act as a trigger for expression of emotional pattern. There is a correlation between characteristic pattern, its context and type of emotion; for instance an act which is at odd with characteristic pattern of ``morality'' creates a significant delta in harmonic value, therefore causes expression of characteristics pattern which is associated with ``guilt''\footnote{Emotional pattern employ inhibitor characteristics to suppress other patterns and become dominant}.   Context plays a major role in\footnote{Through transforming the granularity of expression pattern sensitive to emotional characteristic pattern, or manipulating/changing the response pattern} type of emotion that are expressed\footnote{Emotion expressed as observable referential characteristic} and experienced.

\noindent \textbf{Reason}

\noindent Logical relation model defines the arrangement, sequence, relation and role\footnote{Characteristic role can be as`` expansion activator'', ``facilitator'' or ``inhibitor''} of characteristics according to a reference patterns\footnote{Reference patterns are constructed from characteristics manifested by intrinsic or virtual compositions.}. Reason is based on the ability to observe the logical relation of patterns and construct a \textbf{\textit{referential}} logical relation model\footnote{Based on the ability to Observe, memorize and reference relation of characteristic patterns to each other, expression context and environment.} accordingly. 

\noindent \textbf{Rationality}

\noindent Reason is the ability to infer causality based on observation and construction of the referential logical relation model; rationality is the ability to optimize the relation model and system response cycles\footnote{Rationality involves examining relationship of different patterns (characteristic models) from causality perspective in different context (simulating scenarios)} according to the \textit{context}. It involves using simulation, observation and transformation to create elaborate permutation of expansion scenarios as well as to optimize granularity of the response pattern for the given context. 

\noindent \textbf{Language}

\noindent Spoken language is based on series of auditory patterns; consisting of sound characteristic\footnote{Linguistic characteristics are referential}, expressed and observed by the auditory system to form a referential characteristic model to represent familiar sensation or cognitive patterns\footnote{Auditory patterns acts as the auditory characteristic of the underlying subject of the expression}. Written symbols (characters, words) are referential characteristics based on further abstraction of the auditory patterns.

\noindent \textbf{Conclusion}

\noindent Harmonic theory provides a mathematical framework to describe the structure, behavior, evolution and emergence of harmonic systems. Using the Harmonic theory, the emergence of ad-hoc collaborative systems can be modeled and explained. It also provides a model to describe multi context\footnote{Multi context complex harmonic system can be modeled as harmonic cube} complex systems. Harmonic theory can be used to build new classes of systems, which understand the problem context and find alternative and more efficient solutions\footnote{Efficient solution can be found using Through transformation of either the characteristic model or the underlying subject}, by continuously building and improving their vocabulary of transformation patterns, such systems can increase their efficiency and accuracy within their operating environment.  Such systems would be more effective in understanding, adapting and exchanging with their environment and partners as well as solving unanticipated problems.

\noindent Harmonic theory provides a framework for building Intelligent contextual and adaptive computing, cognition and sensory system that comprise of but not limited to auditory, visual and tactile prime-able and virtualize-able memory and sensory system, it also embodies cognitive apparatus comprising of cluster of harmonic systems that participate in exchange and trade to enhance harmonic status (by gaining access to composition capable of manifesting characteristics that better conform to its harmonic expression). This framework enables the system to:

\begin{enumerate}
\item  Form virtual composition that is able to manifest context sensitive characteristics according to sensory and cognitive system demand.

\item  Detect, observe, memorize and reproduce referential characteristics manifested by compositions in its environment, understand and construct relational logical model of characteristics and develop virtual composition capable of manifesting characteristics according to context and system state. 

\item  Develop a sense of ``self'' through constructing virtual compositions based on referential characteristics deriving from its own intrinsic or external observed compositions.

\item  Simulate virtual scenarios based on imaginary referential characteristics to reduce risk and create new characteristic permutation and arrangements.

\item  Rearrange characteristics of environment, context, composition or expression in order to increase system benefit (harmonic state)

\item  Communicate using abstract symbolized referential characteristics derived from intrinsic and observed compositions that are agreed upon through exchange with peers.
\end{enumerate}

\noindent

\nocite{*}
\bibliography{TheHarmonicTheory}
\bibliographystyle{plain}

\noindent 

\noindent

\end{document}